# Micro-interventions in urban transport from pattern discovery on the flow of passengers and on the bus network

Carlos Caminha, Vasco Furtado, Vládia Pinheiro e Caio Ponte
Programa de Pós-graduação em Informática Aplicada (PPGIA)
Universidade de Fortaleza (Unifor)
Brazil

*Abstract*— *In this paper, we describe a case study in a big metropolis, in which from data collected by digital sensors, we tried to understand mobility patterns of persons using buses and how this can generate knowledge to suggest interventions that are applied incrementally into the transportation network in use. We have first estimated an Origin-Destination matrix of buses users from datasets about the ticket validation and GPS positioning of buses. Then we represent the supply of buses with their routes through bus stops as a complex network, which allowed us to understand the bottlenecks of the current scenario and, in particular, applying community discovery techniques, to identify clusters that the service supply infrastructure has. Finally, from the superimposing of the flow of people represented in the Origin-Destination matrix in the supply network, we exemplify how micro-interventions can be prospected by means of an example of the introduction of express routes.*

## I. INTRODUCTION

A smart city is known through the ownership of data and information produced on it to generate knowledge, which, applied, promotes the well being of citizens. The era of digital information is characterized by the huge capacity of the cities of sensing itself through the data collection on its own dynamic. Within this context, the biggest challenge is to, from this diverse and the voluminous data, to create means to generate knowledge and apply it with effectiveness for the benefit of the collectivity.

One of the areas that best exemplifies the context described above is the urban mobility. The most diverse sensors and digital media record daily information on tracks of people, which becomes a rich input for the carrying out of studies that support public transport policies. In this article we describe our experience in a large Brazilian metropolis within the framework of its project for Smart City. We will show how, from data collected by digital sensors, we tried to understand mobility patterns of persons using buses and how this can generate knowledge to suggest interventions that are applied incrementally into the transportation network in use.

Our contributions are concentrated on three aspects. First, we show that, although there is a great variety and volume of collected data, it is necessary a preparation and refinement work of the data so that they can be used in prospective studies. For this we have developed an algorithm to estimate the Origin-Destination matrix (ODM) of buses users from datasets on ticket validation and on GPS positioning of buses. We show that the sample of trips with origin and destination estimated by our algorithm is statistically significant of the global behavior of bus users. Thus we ended up producing an ODM with quantities well superior to those built by survey that are still the ones that most naturally permeate the work in the field of transport engineering [1].

From this estimate, we chained in our second contribution, which consists of representing the scenario of urban mobility (by making use of the ODM) as complex networks. We built a complex network that represents the supply of public transportation in the city considering the vehicles used, the routes that they follow and the itinerary of the routes with their bus stops. This network has allowed us to understand the bottlenecks of the current scenario and, in particular, applying community discovery techniques, to identify clusters that the service supply infrastructure has.

Finally, our third contribution consisted of superimposing to the network of supplies the flow of people represented in the ODM. This allowed in principle the extension of the characterization of the transport system, emphasizing the flow of people inter and intra community as well as has allowed simulating micro-interventions at the current supply network. Micro-interventions or urban acupuncture [2] advocates the introduction of small modifications in urban scenario so that, with low cost, it is possible to change a local reality and bring impacts on the organism as a whole (in the case, the city). Our approach is proved to be useful in this scenario because it allows not only to intervene locally but also it gives us a chance to have a global view of the impact of this intervention in the transport network as a whole. We provide examples of how this can be done through the introduction of express routes, which can lead to improved public service by increasing the convenience and agility at the pathway of frequent users of an ODM through improvement of global metrics of the network.

## II. Related Work

Recently a growing body of literature has been generated describing the complex networks modeling in context of air [3], rail [4] or urban bus [5] and bikes [6] transport. In general, these studies aim to characterize the networks using metrics such as the degree distribution, the average path length and the clustering coefficient. It is noteworthy mentioning the work of [7] who compared the public transport system in 22 towns in Poland. Among other characteristics, the authors showed that the degree distribution follows a Power Law, in particular, a Small-World one that are hierarchically organized.

The work presented in [8] is particularly relevant in this context and is complementary to our. Unlike the majority of researchers who aim to prevent or reduce overload in the links, the authors developed a heuristic algorithm to optimize the transport network by making traffic balancing through the minimization of maximum node betweenness with the smallest possible path length size, which is particularly useful to avoid congestion on the network due to congestion in a node. Using this strategy, they show that the network can support more traffic without congestion than that in the case when the route is shorter. This is, however, a work, which focuses exclusively on the network representing the infrastructure of public transport.

Regarding the ODM estimation, traditional methods are through large-scale sampled surveys, conducted once in every 1-2 decade. Two main disadvantages of these methods are the financial cost and obsolescence. Alternatively, models using the traffic count on a set of links have been proposed. The accuracy of these models depends on estimation model used, the input data errors and on the set of links with collected traffic counts [9]. In this work we are interested in static ODM for long-time transportation planning. The Entropy Maximization (EM) and Information Minimization (IM) models, and statistical approaches like Maximum Likelihood (ML), Generalized Least Squares (GLS) and Bayesian Inference (BI) have been mostly adopted. However, [9] highlight that few realistic approaches focused on large size network applications. More recently, Artificial Neural Networks for ODM estimation was applied on hypothetical networks with several constraints [10], and more studies still need to be carried out for planning and designing purpose for large size networks.

## III. Representing the Public Transport Network through a Complex Network

### A. Data about Public Transport

For the realization of this research five datasets were used related to network of public transport of Fortaleza in Brazil. The data refer to March 2015 and comprise the following: bus stops; routes; Terminals; bus GPS; and ticket validation. The bus stops are places where passengers go on and go off the vehicles. In all, this data set has 4783 geo-referenced records. Bus routes have 359 records, which have information about the route direction and its itinerary. Bus terminals, the smallest of the data sets, as they possess only seven records, a terminal is a place designed to provide a better service in the user transshipment, various bus routes pass through terminals. The bus GPS dataset data has around 104 million records. The buses register their location every thirty seconds. In total 2034 buses circulated in Fortaleza in March 2015.

The user of bus in Fortaleza can use a smart card as a ticket for validating a trip. With the smart card it is possible to make connections in any point of the city as long as it is within the period of two hours since the last ticket validation. A validation occurs when the user passes his or her smart card in the bus or in a terminal validator.

The dataset of ticket validation refers to the month of March and has about 29 million recordings made in bus or directly at the validator of a bus terminal. The basic data are the user´s id, data and time of the validation, bus route identification, terminal identification (in case of validation in a terminal validator), and vehicle identification.

Figure 1 illustrates the distribution of validations made in the days of March 2015. There is a pattern for days of the week, another for Saturdays ($7^{th}$, $14^{th}$, $21^{st}$, $28^{th}$) and another for Sundays ($1^{st}$, $8^{th}$, $15^{th}$, $22^{nd}$) and public holidays (the $19^{th}$ and $25^{th}$). Practically all the days of the week that were not holidays there were approximately 1.1 million validations made, the only exceptions were the days 5 and 6, while the first there were 2.3 million validations and the second had only 838. The administrator of the database believes that there must have been some errors generating the file, for this reason, these data will be disregarded in our analysis.

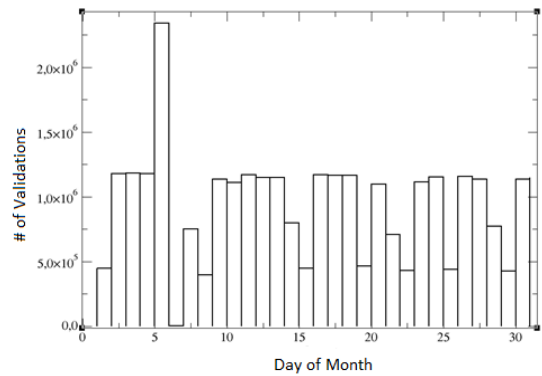

*Figure 1. Ticket validations in March*

### B. Generating the Origin-Destination Matrix

To perform the origins estimates and the user's destinations, the data already detailed in Section III were used. The problem that we faced was that the raw data did not allow saying precisely the bus stop where the passenger got on the bus, because they contain the time that a given user validated his ticket on the bust ratchet. We had first to estimate the place where this validation was done. For this reason, we retrieved in the bus GPS dataset the latitude and the longitude where the bus was at that validation time. However, this geographical coordinate is not necessarily the origin of the passenger, since he/she may not have done the validation of the ticket at the time that got on the bus. He could, for example, have remained at the rear of the vehicle up to the moment in which the bus approached from its destination. This problem can be minimized as we detail further.

With regard to the destination of the user unfortunately there is no precise record that indicates where the passenger got off. However, in literature, we found some heuristics that aim at supporting the estimate of that point of going off the bus with good reliability. The first is that, in their last journey in the day, the majority of people tend to return to the point where they set off [11, 12]. That is, in a day where the passenger has made only two trips, for example, the location of the first validation corresponds to the origin and the location of the second validation has good probability of being the destination. With this you can assume with certain reliability that the pair origin and destination (OD pair) of this user in question corresponds to the ticket validation coordinates.

The OD pairs may still be generalized to the cases where the passengers take more than two trips in a day. If the user has made three trips in a day, you can assume that the first trip corresponds to his first origin at the day, O1, the second trip has at its validation point at the first destination, D1, and the second origin of the day, O2. The third journey, in its turn, has in its Single ticket validation location at the second destination at the day, D2, and the third the origin, O3. Closing the triangle of OD pairs, it is assumed that the user will terminate their paths of the day returning to the point of first origin in the day, which in this case would also be the destination, D3, the user. The OD pairs in this example would then be {O1, D1}, {O2, D2} and {O3, D3} where in terms of geographical coordinates D1 = O2, D2 = O3 and O1 = D3. This example exposed in this graph may be associated with the case of a citizen who, for example, leaves the house in the morning toward the location where he or she works, then in another shift, goes to the university and then back home at the end of their daily routine.

In [11, 12] different strategies have been proposed for the validation of origin and destination. One of them says that the behavior of a recurring user must be checked. In days equal in the week, it is expected that the user leaves the same origins to the same destinations, for example, if in almost every Monday of a month, a user "X" leaves a point P1 for a P2 and then to a point P3, it should be assumed that this is the default for moving the user "X" on Mondays. Then if in few Mondays it is recorded that he made only the route P1 to P3, it must be verified if these origins and destinations are correct. In this work it is verified if the route taken by the user from P1 arrives in P3, if not reached, it is verified if the same arrives in P2, and if he or she does, then it is added to the user analyzed the point P2 as an intermediary point in the itinerary. In case it is not possible to arrive in P2 from the bus route of P1, the OD is removed. This user scenario described usually occurs because possibly the user made part of the path that day by another modal such as bicycle, train or subway.

As it has already been mentioned, the point where the user made the ticket validation may not necessarily be the origin of the same, because this user may not have validated his or her ticket at the time that he or she got on the bus. The literature suggests [11,12] that to give more precision to these points, it must also be adopted a strategy to analyze the behavior of a recurring user. A person normally sets off from the same origins, but may have validated the ticket in different points of the bus route (capacity of the vehicle can motivate this), then it must be adjusted all the origins of the user in question to the first point of the validation found in the days analyzed. This is probably the closest point of origin of the user.

Using the strategies outlined in this section, the OD pairs have been generated of the bus network of Fortaleza for a week in the month of March 2015. The chosen period was from 11th (Wednesday) today 17th (Tuesday). This week has been chosen due to the absence of holidays and also due to the fact that the same does not possess the inconsistencies mentioned in section III of this work. In total, around 1.7 million OD pairs have been generated in this period.

*C. Statistical validation of the representativeness of the OD sample spatial*

Typically, ODM are constructed via surveys with a sample of the users. It is expected that the samples guarantee, with a certain degree of reliability, that the ODs estimated are correct. A major challenge is to ensure that the sample is not biased spatially, for example, that the embarkings into the bus used in estimates are not proportional (considering each bus stop) to the total of embarkings.

To spatially validate the sample we have generated we have checked the correlation between the embarkings used to estimate the OD pairs and the total of embarkings in the bus stops in Fortaleza. From the equation $y = Y * x^\beta$, where $x$ represents the total of embarkings at the bus stops, $y$ the total of embarkings used at the estimates and $Y$ is a normalization constant, we say that a linear or isometric relationship, as evidenced by the value of the exponent $\beta$ close to one, indicates that, proportionally, the embarkings used in estimates are equivalent to the total of embarkings in bus stops, noting that the samples were extracted in random order.

Figure 2 shows the correlation between the sample of the embarkings used in the estimation of origins and destinations and the total of bus embarkings recorded in the period studied. The red line represents the regression made about the data, it was estimated the exponent $\beta = 0.95 \pm 0.02$ with $\beta = 0.95 \pm 0.02$ and $R^2 = 0.92$. Also the confidence intervals of the correlation were estimated for the data distribution $\{X_i, Y_i\}$. It was applied the method Nadaraya-Watson [13] to build the kernel smoother function,

$$m_h(x) = \frac{\sum_{i=1}^{N} K_h(x - X_i) Y_i}{\sum_{i=1}^{N} K_h(x - X_i)},$$

Where *n* is the number of points of distribution and $K_h(x - X_i)$ is a Gaussian kernel function formally defined as $K_h(x - X_i) = \exp\left[\frac{(x - X_i)^2}{2h^2}\right]$, where $h$ is the estimate of bandwidth by minimum squares using cross-validation [13,14]. We compute the 95% confidence interval (CI) 500 random bootstrapping samples with replacement.

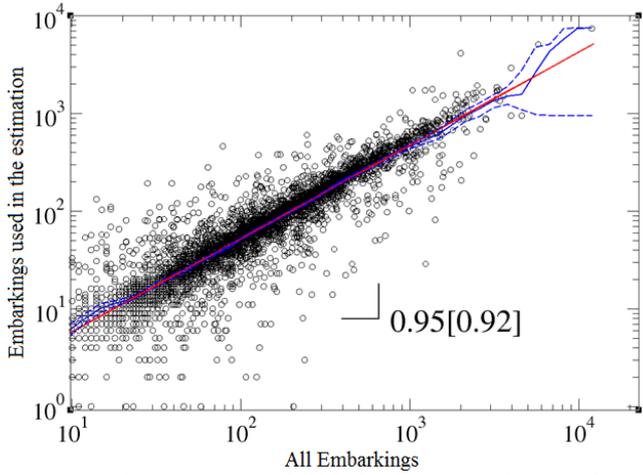

*Figure 2. The isometric relationship between the sample and the total of embarkings.*

IV. CHARACTERIZATION OF THE SUPPLY NETWORK

*A. Public Transport Network*

The transport infrastructure network of buses of a city is determinant in the movement of persons making the same reach their places of desire for performing daily activities. In the specific case of bus network, the nodes are bus stops and the bus routes that lead people between these bus stops are the edges. The itinerary of a route is the set of bus stops visited during a route. It is said that a route finished its itinerary when the bus reaches its last stop.

We can thus model a network as a directed graph, $G(V, E)$, with vertices, $v (\in V)$, representing the bus stops and edges, $e (\in E)$, supplies between two bus stops. Formally the weight of the edge, $w_{v_i \to v_j}$, represents the bus supply between two bus stops $v_i$ and $v_j (\in V)$. This supply is then calculated from the summing up of the bus routes weight, $w_{Li}$, that pass by two bus stops. Formally $w_{v_i \to v_j} = \sum_{i=1}^{N} w_{Li}$, where $N$ is the total number of routes that visit, in sequence, $v_i$ and $v_j$. In its turn, the weight of the bus routes, is calculated from the product of the quantity of vehicles, $V_i$, by the number of times that a vehicle of this route finishes its itinerary in a day, $C_i$. That is, $w_{Li} = V_i C_i$.

The concept of "completing the itinerary", done by a route, is fundamental in this calculation because the itineraries can have different sizes, a route that has more vehicles than another not necessarily has a higher supply. Said another way, the greater the weight of $w_{Li}$, the more the vehicles of $L_i$ pass by their itinerary´s bus stops.

*B. Detection of communities*

From the data described in Section III.A we generated a network with 4.783 vertices and 6.103 edges representing the infrastructure of the Fortaleza bus. In total this graph has 10 connected components. The giant component contains 4.761 nodes, precisely 99,35% of the total of nodes. Due to the small relevance of the minor components, it will be conducted the rest of the study using only the giant component.

We have investigated also if the network (giant component) could be divided into sub nets with high internal clustering coefficient and low connectivity with external components. This might indicate the existence of areas where the bus supply is privileged and also bottlenecks (weak links) between these areas. It was used the algorithm [15] to detect the communities. This algorithm makes use of a heuristic method based on the optimization of modularity of graphs. The modularity of a set of nodes is measured by an actual value between 0 and 1, being computed by the ratio between the quantity of edges that connect the elements of the set among themselves by the total of the edges of the set of nodes [15-17]. The algorithm can be divided in two phases that are repeated iteratively. In the first phase it is assigned a different community for each network node. Thus, in this initial split there are many communities as nodes. For each node $i$ are considered the neighbors of $i$ and $j$ is evaluated the gain of modularity that occurs when moving $i$ of their community to the community of $j$. The node $i$ is then placed in the Community for which this gain is maximum, but only if this gain is positive. If no positive gain is possible, $i$ remains in their original community. This process is applied to all nodes repeatedly until no improvement can be achieved and the first phase is thus completed. The second phase of the algorithm consists of the construction of a new network whose nodes are the communities found during the first phase. To do this, the weights of the connections between the new nodes are given by the sum of the weight of the connections between the nodes in the two corresponding communities. The connections between the nodes of the same community are represented by self-loops for this node of the new network. The phases are repeated until the maximum global modularity is found.

V. EXPERIMENTAL EVALUATION

Using [15], ten communities were found in the bus network of Fortaleza. The communities of the bus network of Fortaleza have modularity 0.877. Table I shows information about the structure of the communities; it illustrates for each community its identifier, number of nodes, diameter, average clustering coefficient and weighted average degree. Since the diameter is a property that is very sensitive to the size of the network, it was added to it a standardized value since the communities are not balanced by the quantity of nodes. This normalized value was calculated by dividing the diameter of each community by their quantity of nodes.

In the context of the proposed network, a region or community to meet well the population should maximize their average clustering coefficient. This is because in that network the semantics of the edges connection is the existence of bus routes following these paths. In other words, a higher clustering coefficient directly implies in greater possibilities of locomotion within the region. In this indicative of mobility, highlight are the communities two, four and seven, each with average coefficient greater than 0.015, a value way greater than that of the community one for example.

Another hint of internal mobility is the diameter of the net. The diameter in a community is the distance of the shortest path between the two most distant vertices. Communities with greater diameter indicate the existence of at least a difficult path in the region analyzed, but this high value may indicate the existence of more than one difficult path [18]. In Table I it can be noticed that the Communities 0 and 9 stand out having standardized diameters 0.089 and 0.090 respectively. These are the values almost three times smaller than the diameter of the Community 1, for example.

TABLE I. MAIN FEATURES OF THE COMMUNITIES

| Id | Nodes | Diameter | | Density | Avg clustering coefficient | Avg weighted degree |
|---|---|---|---|---|---|---|
| | | Value | Normalized Value | Value | Value | Value |
| 0 | 701 | 63 | 0,089 | 0,002 | 0,010 | 440,539 |
| 1 | 434 | 106 | 0,244 | 0,003 | 0,001 | 600,274 |
| 2 | 366 | 53 | 0,144 | 0,003 | 0,018 | 635,871 |
| 3 | 397 | 62 | 0,156 | 0,003 | 0,012 | 558,622 |
| 4 | 186 | 39 | 0,209 | 0,006 | 0,016 | 609,815 |
| 5 | 572 | 71 | 0,124 | 0,002 | 0,009 | 496,279 |
| 6 | 379 | 58 | 0,153 | 0,003 | 0,007 | 433,505 |
| 7 | 461 | 62 | 0,134 | 0,003 | 0,017 | 584,867 |
| 8 | 590 | 107 | 0,181 | 0,002 | 0,012 | 481,195 |
| 9 | 675 | 61 | 0,090 | 0,002 | 0,008 | 354,368 |

Table I also shows the density of each community. Dense communities indicate a trend toward a better mobility, because the existence of many edges is directly related to a greater supply of routes in the network. We emphasize the Community 4 that has value three times greater than the communities 0, 5, 8 and 9. In the context of the proposed network, a greater weighted degree indicates greater supply of public transport in the region. In this indicative the community that stands out the most is number two and that has the worst index is Community 9.

### A. Inter and Intra communities flow

Being characterized the bus network structure from Fortaleza, it was verified how the communities found relate to the users´ displacements. First, we computed the intra and inter communities moving. Bus stops from the OD pair that are within the same community are accounted for as an intra-Community movement. Otherwise it is recorded as a cross-community (inter-community) movement.

During the days of the week it was verified that 60.2% of the movements are inter-communities and 39.8% are intra-community. On Saturday the inter-community movements decrease to 56.1% and on Sunday to 49.8%. This movement drop, can be explained to the fact that, in the week, the citizens´ obligations, such as work and study, lead them to make longer displacements. On Saturday there is an increase in the intra-community movements, but the percentages are still relatively close to the days of the week, this must be explained to the fact that, on Saturdays, many people still have obligations such as work and study, and with this, there is still need for a longer travel. On Sunday, it occurs the first time at the week where inter-community movements are minority, this probably occurs to the fact that Sunday is a day when these people are in movement for leisure activities, then, to have a greater convenience, they look for closer displacements.

By ranking the communities from the flow of people it was possible to verify that Community 0 has more than twice the flow of the second largest with 28.8% of intra-community flow across the net. This can be explained by the fact that this community has the two largest shopping malls in the city, in addition to large residential and commercial area. Concerning this community, in Section V. A. of this work, it was shown that, among all the properties of the offer of the same, it only had highlight in its diameter, for all the other metrics it showed a value under the other communities. This may indicate problems in relation to the care of demand, since the community with greater flow of people has efficiency proprieties under the others.

### B. Methodology for the creation of micro-interventions on the net

In the previous section it was shown that during the days of the week, moment in which the network is more used, the inter-community flow is the most intense. Detection of communities, applied to the network studied here, allows discovering the city areas where there is a greater supply for the internal flow and difficulty of mobility to other areas. Making use of metrics of complex networks, we have proposed a methodology for creation of express routes that connect the communities with greater inter flow demand. We simulated the creation of edges that represent express routes by connecting the center of communities and assess the impact of these routes in the network structural characteristics.

To find the central node of the communities, the metric betweenness centrality [18, 19] was used. It is calculated as, $\sum \frac{\sigma_{st}(v)}{\sigma_{st}}$ where $\sigma_{st}$ is the number of the shortest paths from $s$ for a node $t$ and $\sigma_{st}(v)$ it the number of these paths which pass by $v$.

The centers of the communities were estimated and to serve as an experiment, five edges connecting the communities with greater inter flow were created. These edges symbolize express routes offering a service where the buses connect the city areas stopping just in the centers of the Communities. It will be examined the effect that micro-interventions have over three structural characteristics that measure the quality of mobility in a complex network: average path length, average eccentricity and diameter.

The interventions were made between the communities that have a greater inter flow. The greatest flow occurs between the communities 0 and 5; daily, 10% of users move between those communities. The other four links with greater flow are among the Communities: 1 and 2; 0 and 1; 5 and 7; and 0 and 7. Figure 3 shows the values of the metrics to the extent that the interventions were created. The x-axis represents the interventions, while the y-axis shows the values of the metrics. The average path length of a network measures the average cost of moving in the network. Figure 3, black line, illustrates that, before the interventions, 38.26 hops were necessary, in average, to reach any point in the net. That is, users of the bus system of Fortaleza must pass by 38 bus stops before reaching their destination. After the creation of the first intervention the network suffers the largest drop in the value of this metric, reaching 27.26. Nine hops fewer on average. Whereas the other interventions are made the length of path falls, but very gradually. On average, a hop by intervention, reaching: 24.25 after the fifth intervention.

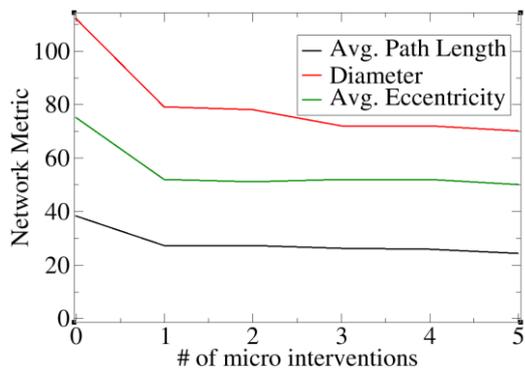

*Figure 3 structural characteristics of the buses network supply after the micro-interventions.*

The average path length takes into account all possible paths within the network to have its value calculated, in practice it is an optimistic measure within the network that is being studied, since it is known that the demand is greater for longer movements, Inter community movements. Thinking about that, it was measured the eccentricity of the bus network nodes from Fortaleza. The eccentricity of a node, $v$, is the length of the shortest path $v$, $v$ up to the most distant node from it [18]. On average the eccentricity of the original network is 75.43 hops (green line in Figure 3). After the first intervention the average eccentricity is reduced to 52.34 hops and, again, the other interventions have a minor impact on the metric that, after the fifth intervention, reaches 50.01 hops.

Figure 3 also shows the behavior of the diameter of the network (red line). The diameter may be defined as the largest eccentricity of a network [18,19]. Before the interventions its value was 112 hops, after the creation of the first micro-intervention its value is reduced to 79, reaching 70 hops after the creation of the fifth. This behavior shown by the metrics exposed in Figure 3 shows how refined this information is for an urban planner. It would be very difficult to identify a small intervention in the network structure, which impacted effectively in the movement of persons, without using exploratory methods exposed here.

## VI. Conclusion

This work explored the supply and demand for bus networks in a case study with a large Brazilian metropolis. Models and metrics of complex networks have been used to understand the dynamics of urban mobility, which led us to describe a methodology to suggest micro-interventions from the analysis of the communities found in the bus supply networks. This methodology is directed to the government and is therefore useful for allowing incremental and gradual interventions, which is desirable in situations in which caution is advised in the actions, either by the complexity of the consequences that the intervention may cause, either by budget limitations that many sometimes it imposes.

Secondary contributions of this work are, first, the reconstruction of the origins and destinations of users from public data bus GPS and validation of single ticket. Second, the characterization of the offer of the city's bus network making use of complex networks, where, among other things, it was discovered that the network has bottlenecks that hinder mobility certain parts of the city. Third, the characterization of the network demand, which analyzed the flow of people from the perspective of intra and inter communities, was exposed that network bottlenecks were overwhelmed by inter-community flow, more intense, especially on weekdays.